\documentclass[conference]{IEEEtran}

\ifCLASSINFOpdf
  \usepackage[pdftex]{graphicx}
  \graphicspath{ {pictures/}{../pdf/}{../jpeg/} }
  \DeclareGraphicsExtensions{.pdf,.jpeg,.png,.PNG,.svg,.SVG}
\else
\fi
\ifCLASSOPTIONcompsoc
 \usepackage[caption=false,font=normalsize,labelfont=sf,textfont=sf]{subfig}
\else
 \usepackage[caption=false,font=footnotesize]{subfig}
\fi
\usepackage{flushend}
\usepackage{tabularx,booktabs,textcomp}
\newcolumntype{C}{>{\centering\arraybackslash}X} 
\newcolumntype{L}{>{\raggedright\arraybackslash}X} 
\usepackage[noadjust]{cite}

\usepackage{wasysym}
\usepackage{lipsum} 
\usepackage{svg}
\usepackage{amsmath}
\usepackage{amssymb}
\usepackage{multirow}
\usepackage[none]{hyphenat}

\begin{document}

\title{Residual Transformer Fusion Network for Salt and Pepper Image Denoising}

\author{
\IEEEauthorblockN{Bintang Pradana Erlangga Putra}
\IEEEauthorblockA{Department of Informatics\\
Universitas Sebelas Maret\\
Surakarta, Indonesia\\
bintangpradana02@student.uns.ac.id}
\and
\IEEEauthorblockN{Heri Prasetyo}
\IEEEauthorblockA{Department of Informatics\\
Universitas Sebelas Maret\\
Surakarta, Indonesia\\
heri.prasetyo@staff.uns.ac.id}
\and
\IEEEauthorblockN{Esti Suryani}
\IEEEauthorblockA{Department of Informatics\\
Universitas Sebelas Maret\\
Surakarta, Indonesia\\
estisuryani@staff.uns.ac.id}
}

\maketitle

\begin{abstract}

Convolutional Neural Network (CNN) has been widely used in unstructured datasets, one of which is image denoising. Image denoising is a noisy image reconstruction process that aims to reduce additional noise that occurs from the noisy image with various strategies. Image denoising has a problem, namely that some image denoising methods require some prior knowledge of information about noise. To overcome this problem, a combined architecture of Convolutional Vision Transformer (CvT) and Residual Networks (ResNet) is used which is called the Residual Transformer Fusion Network (RTF-Net). In general, the process in this architecture can be divided into two parts, Noise Suppression Network (NSN) and Structure Enhancement Network (SEN). Residual Block is used in the Noise Suppression Network and is used to learn the noise map in the image, while the CvT is used in the Structure Enhancement Network and is used to learn the details that need to be added to the image processed by the Noise Suppression Network. The model was trained using the DIV2K Training Set dataset, and validation using the DIV2K Validation Set. After doing the training, the model was tested using Lena, Bridge, Pepper, and BSD300 images with noise levels ranging from 30\%, 50\%, and 70\% and the PSNR results were compared with the DBA, NASNLM, PARIGI, NLSF, NLSF-MLP and NLSF-CNN methods. The test results show that the proposed method is superior in all cases except for Pepper's image with a noise level of 30\%, where NLSF-CNN is superior with a PSNR value of 32.99 dB, while the proposed method gets a PSNR value of 31.70 dB.

\end{abstract}


\section{Introduction}

\IEEEPARstart{C}{onvolutional} Neural Network (CNN) has been widely used to process unstructured datasets such as image classification, image segmentation, and image generation \cite{chen2018encoderdecoder, howard2017mobilenets, radford2016unsupervisedrepresentationlearningdeep}. The utilization of Convolutional Neural Network (CNN) has proven that this method is reliable in computing spatial information such as digital image processing.

The use of transformer architecture is growing in tandem with the development of the use of convolutional architecture in image processing with the Convolutional Neural Network (CNN) method. Nonetheless, the transformer method is still more widely used in Natural Language Processing (NLP) \cite{vaswani2017attention}. The Vision Transformer (ViT) method, which was introduced as a pure transformer architecture method, demonstrates that transformer architecture can also be used in image classification problems \cite{dosovitskiy2020vit}. Convolutional Neural Network (CNN) was introduced to the Vision Transformer (ViT) to form a Convolutional Vision Transformer (CvT) which combines the advantages of convolution and transformer architectures so that it is expected to provide more optimal results in computational efficiency of image processing \cite{wu2021cvt}.

Image denoising has a problem, namely, some image denoising methods require some prior knowledge of information about noise, such as noise model, noise distribution, and noise level, to reduce noise in the image \cite{prayuda2021awgn}. To overcome these problems, Convolutional Vision Transformer (CvT) and Residual Networks (ResNet) are used by implementing end-to-end training so that the Convolutional Vision Transformer (CvT) model can directly map noisy images with clean images.

\section{Proposed Denoising Model}
\label{sec:op}

\begin{figure*}[!t] 
  \centering
  \includegraphics[width=\textwidth]{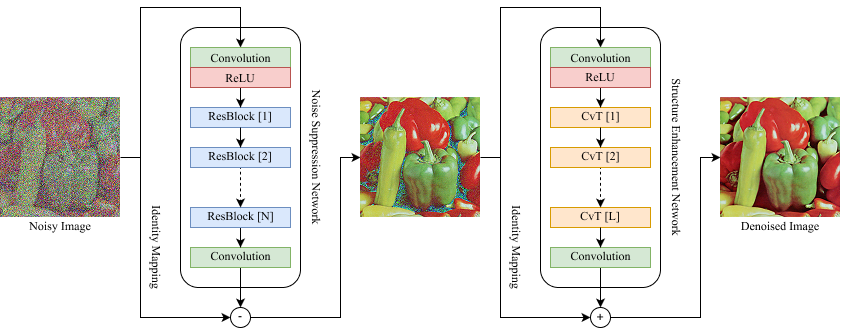}\\
  \caption{RTF-Net Architecture}
  \label{fig:rtf-net}
\end{figure*}

The proposed denoising architecture will be explained in this section, which includes the attention mechanism, transformer, and Convolutional Vision Transformer (CvT). This section will also explain salt and pepper noise modeling and how the proposed architecture eliminates salt and pepper noise from images.

\subsection{Noise Modelling} 

Image denoising is a noisy image reconstruction process that aims to reduce additional noise from the image with various strategies. Some image denoising methods require several prior knowledge of noise information, such as noise model, noise distribution, and noise level, to reduce noise in the image. In an ideal situation, the image denoising algorithm should be able to reduce noise in the image directly without any prior information \cite{prayuda2021awgn}.

\begin{equation}
    \label{eq:noise-model}
    I(i,j)=
    \begin{cases}
    0,&  r_1 < p \land r_2 < 0.5 \\
    255,& r_1 < p \land r_2 \geq{0.5} \\
    I(i,j),& r_1 \geq{p}
    \end{cases}
\end{equation}

In this study, the digital image used will be added with salt and pepper noise where the value of a pixel becomes maximum or minimum depending on a predetermined probability. Equation (\ref{eq:noise-model}) is the formulation of salt and pepper noise with $p \in (0,1)$ as the noise level, $r_1$ and $r_2$ as two pseudo-random values generated at each pixel in the image to be affixed with salt and pepper noise.

\subsection{Residual Transformer Fusion Network}

In this research, an architecture called Residual Transformer Fusion Network (RTF-Net) is proposed and shown in Figure \ref{fig:rtf-net}. In general, the processes that occur in this architecture can be divided into two parts, Noise Suppression Network (NSN) and Structure Enhancement Network (SEN). Residual Block is used in the Noise Suppression Network and is used to learn the noise map in the image, while the Convolutional Vision Transformer (CvT) is used in the Structure Enhancement Network and is used to learn the details that need to be added to the image that has been processed by the Noise Suppression Network. The RTF-Net architecture can be formulated as equation (\ref{eq:nsn}) and equation (\ref{eq:sen}).

\begin{equation}
    \label{eq:nsn}
    y=\tilde{I}-\mathbb{NSN}(\tilde{I};\theta)
\end{equation}

In the first stage formulated by equation (\ref{eq:nsn}), the noisy image denoted as $\tilde{I}$ will be processed using a Noise Suppression Network denoted as $\mathbb{NSN}(\cdot;\theta)$ with $\theta$ as the parameter, then the result will be subtracted by the noisy image and produce an image $y$ which is transition image.

\begin{equation}
    \label{eq:sen}
    I^{*}=y+\mathbb{SEN}(y;\theta)
\end{equation}

In the second stage formulated by equation (\ref{eq:sen}), the $y$ mage will be processed using the Structure Enhancement Network which is denoted as $\mathbb{SEN}(\cdot;\theta)$ with $\theta$ as the parameter, then the results will be summed with the $y$ image to produce a clean image which is denoted as $I^{*}$.

\subsection{Residual Block}

\begin{figure}[h]
    \centering
    \includegraphics[width=2.5in]{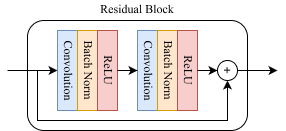}
    \caption{Residual Block Architecture}
    \label{fig:resblock}
\end{figure}

Residual Block architecture is used in the Noise Suppression Network on RT-Net and is shown in Figure  \ref{fig:resblock}. This residual block consists of two convolution blocks, where the convolution block consists of a convolution layer, a normalization function layer, and a ReLU activation function layer. Incoming features will go through layers in the convolution block and then the features that do not pass through the convolution block are summed with the features that pass through the convolution block (identity mapping \cite{he2015residual}) to form output features.

\begin{equation}
    \label{eq:convblock}
    x_{out}=\delta(\aleph(x_{in}*W+b))
\end{equation}

The convolution block can be formulated as equation (\ref{eq:convblock}) where $x_{in}$ represents the input feature, $x_{out}$ represents the output feature, $W$ represents the weight of the convolution layer, $b$ represents the bias of the convolution layer, $\aleph(\cdot)$ represents the normalization function, and $\delta(\cdot)$ represents the activation function.

\begin{equation}
    \label{eq:resblock}
    x_{out}=x_{in}+\mathbb{B}(\mathbb{B}(x_{in};\theta);\theta)
\end{equation}

The residual block can be formulated as equation (\ref{eq:resblock}) where $x_{in}$ represents the input feature, $x_{out}$ represents the output feature, and $\mathbb{B}(\cdot;\theta)$ represents the convolution block and its parameters.

\subsection{Convolutional Vision Transformer}

\begin{figure}[h]
    \centering
    \includegraphics[width=2.5in]{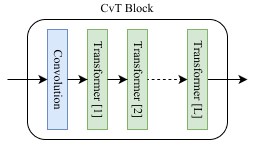}
    \caption{CvT Architecture}
    \label{fig:cvt}
\end{figure}

The Convolutional Vision Transformer (CvT) \cite{wu2021cvt} depicted in Figure \ref{fig:cvt} is a hybrid of convolutional and Vision Transformer (ViT) architectures. The goal of this method is to achieve the advantages of the Convolutional Neural Network (CNN) in the domains of local receptivity, shared weights, and spatial subsampling with the advantages of transformers in dynamic attention, global context fusion, and improved generalization.

\begin{equation}
    \label{eq:cvt}
    x_{out}=\mathbb{T}(x_{in}*W_{in}+b_{in};\theta)
\end{equation}

Convolutional Vision Transformer architecture can be formulated as equation (\ref{eq:cvt}), where $x_{in}$ represents the input feature, $W_{in}$ and $b_in$ represent the weight and bias of the convolutional embedding layer, and $\mathbb{T}(\cdot;\theta)$ represents the transformer layer and the parameters used in that layer.

\begin{figure}[h]
    \centering
    \includegraphics[width=2.5in]{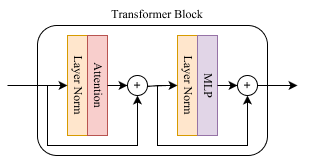}
    \caption{Transformer Architecture}
    \label{fig:transformer}
\end{figure}

Figure \ref{fig:transformer} depicts the Transformer Block architecture design utilized in the Convolutional Vision Transformer (CvT) block. A transformer Block normally consists of two components. First, the features are input and processed in the first component, namely the normalization function layer and the attention layer, which is then summed to the identity mapping of the incoming features which is defined as equation (\ref{eq:transformer1}). Following the first part, the features will be processed in the second part, which comprises of a normalizing function layer and a Multi-layer Perceptron (MLP) layer, followed by an identity mapping of the features that enter part two, which is expressed as an equation (\ref{eq:transformer2}).

\begin{equation}
    \label{eq:transformer1}
    x^*=x_{emb}+\mathbb{A}(\aleph(x_{emb}))
\end{equation}

When the embedding results are processed by the transformer, the feature will pass through the normalization layer, the attention mechanism, and the embedding results to form a residual connection. In equation (\ref{eq:transformer1}), $x_{emb}$ represents the embedding result, $\aleph$ represents the normalization layer, $\mathbb{A}$ represents attention mechanism, and $x^*$ represents the normalization, attention, and residual connection results.

\begin{equation}
    \label{eq:transformer2}
    x_{out}=x^*+\mathbb{M}(\aleph(x^*))
\end{equation}

The result of equation (\ref{eq:transformer1}) will then be processed through the normalization layer, multi-layer perceptron, and residual connection. In equation (\ref{eq:transformer2}), $x^*$ represents the calculation result of equation (\ref{eq:transformer1}), $\aleph$ represents the normalization layer, $\mathbb{M}$ represents the multi-layer perceptron formulated by equation (\ref{eq:mlp}), and $x_{out}$ represents the normalized, multi-layer perceptron, and residual connection results.

\begin{equation}
    \label{eq:mlp}
    x_{out}=\delta(x_{in}*W_1+b_1)*W_2+b_2
\end{equation}

Equation (\ref{eq:mlp}) is the formulation of the multi-layer perceptron architecture used in equation (\ref{eq:transformer2}). This architecture consists of three layers namely, the convolution layer, gaussian error linear unit (GELU), and convolution layer \cite{hendrycks2016gelu}. In equation (\ref{eq:mlp}), $x_{in}$ represents the input feature, $x_{out}$ represents the output feature, $\delta$ represents the activation function, $W$ represents the weight of the convolution layer, and $b$ represents the bias of the convolution layer.

\subsection{Attention Mechanism}

\begin{figure}[h]
    \centering
    \includegraphics[width=3.5in]{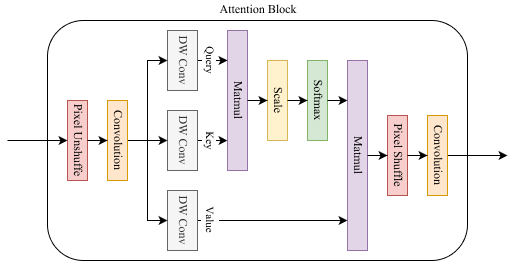}
    \caption{Attention Architecture}
    \label{fig:attention}
\end{figure}

We employ Multi-head Attention \cite{vaswani2017attention} with Attention Block in each Transformer Block, as illustrated in Figure \ref{fig:attention}. The use of direct attention can result in excessive computational load. To overcome this problem, we use pixel unshuffle to reduce the spatial dimensions of the image and move it to channel dimensions in the image, convolution is then applied to the unshuffled pixel image to reduce the channel dimensions in the image.

\begin{equation}
    \label{eq:attention1}
    x_{red}=\mathbb{S}^{-1}(x_{in})*W_{red}+b_{red}
\end{equation}

The above process can be formulated as equation (\ref{eq:attention1}), where $W_{red}$ represents the convolution weights, $b_{red}$ represents the convolution bias, $\mathbb{S}^{-1}(\cdot)$ represents the pixel unshuffle operator, $x_{in}$ represents the input feature, and $x_{red}$ represents the output feature. The features resulting from the above process will then enter the Multi-head Attention mechanism.

\begin{equation}
    \label{eq:attention2}
    \hat{x}_{attn}^{(h)}=\mathbb{S}(x_{attn}^{(h)})*W_{proj}
\end{equation}

The next stage is pixel shuffle where the channel dimensions are converted into spatial dimensions and then projected using a convolution layer. The process can be formulated by equation (\ref{eq:attention2}) where $\mathbb{S}(\cdot)$ represents the pixel shuffle operator, $W_{proj}$ represents the weight of the projection layer, $x_{attn}^{(h)}$ represents the incoming feature resulting from the $h$-th attention head processing, and $\hat{x}_{attn}^{(h)}$ represents output features that have been processed by $h$-th attention head.

\subsection{Training}

Experiments were conducted in this paper utilizing the Python programming language and the Pytorch framework. The model was created using a system with the following specifications: AMD Ryzen Threadripper 12 Core 24 Threads, Nvidia RTX 3090 GPU with 24GB VRAM, and 128GB RAM.

\begin{table}[h]
\caption{Hyperparameter}
\label{tab:hyperparameter}
\centering
\begin{tabular}{lc}
\toprule
    \textit{Hyperparameter} &   Nilai\\
\midrule
    \textit{Patch Size}     &   64$\times$64    \\
    \textit{Kernel Size}    &   3$\times$3      \\
    \textit{Stride}         &   1               \\
    \textit{Padding}        &   1               \\
    \textit{NSN Depth}      &   8               \\
    \textit{NSN Features}   &   32              \\
    \textit{SEN Depth}      &   2               \\
    \textit{NSN Features}   &   32              \\
    \textit{CvT Depth}      &   2               \\
    \textit{Attention Head} &   4               \\
    \textit{Optimizer}      &   Adam            \\
    \textit{Learning Rate}  &   0.001           \\
    \textit{Batch Size}     &   32              \\
    \textit{Epoch}          &   25              \\
    \textit{Step Size}      &   6               \\
    \textit{Gamma}          &   0.5             \\
    
\bottomrule
\end{tabular}
\end{table}

At this stage, hyperparameter tuning of the Residual Transformer Fusion Network (RTF-Net) is carried out. The hyperparameters considered include patch size, image size, number of layers on the Noise Suppression Network (NSN), number of layers on the Structure Enhancement Network (SEN), kernel size, number of features, optimizer, learning rate, batch size, and epoch. These hyperparameters are shown in Table \ref{tab:hyperparameter}.

\section{Results and Discussion}

\begin{table*}[!t]
\caption{Peak Signal to Noise Ratio Result}
\label{tab:result}
\centering
\newcolumntype{Y}{>{\centering\arraybackslash}X}
\begin{tabularx}{\textwidth}{@{}lYYYYYYYY@{}} 
\toprule
\multirow{2}{*}{Image}  & \multirow{2}{*}{Noise Level} & \multicolumn{7}{c}{Method}                                                    \\ 
\cmidrule{3-9}
                        &                              & DBA\cite{srinivasan2007dba}   & NASNLM\cite{varghese2015nasnlm} & PARIGI\cite{delon2016parigi} & NLSF\cite{fu2018nlsf}  & NLSF-MLP\cite{burger2012nlsf} & NLSF-CNN\cite{fu2018nlsf}       & Ours            \\ 
\midrule
\multirow{3}{*}{Lena}   & 30\%                         & 34.42 & 28.09  & 33.90  & 34.20 & 30.80    & 35.38          & \textbf{38.87}  \\
                        & 50\%                         & 30.11 & 26.15  & 29.91  & 30.12 & 29.28    & 32.55          & \textbf{34.62}  \\
                        & 70\%                         & 25.84 & 25.97  & 25.22  & 25.79 & 27.63    & 30.18          & \textbf{32.85}  \\ 
\midrule
\multirow{3}{*}{Bridge} & 30\%                         & 28.07 & 23.68  & 25.19  & 28.21 & 25.19    & 28.71          & \textbf{32.24}  \\
                        & 50\%                         & 24.24 & 22.91  & 22.61  & 22.45 & 23.86    & 26.01          & \textbf{28.26}  \\
                        & 70\%                         & 21.21 & 22.63  & 20.06  & 21.02 & 22.61    & 24.11          & \textbf{26.44}  \\ 
\midrule
\multirow{3}{*}{Pepper} & 30\%                         & 26.85 & 22.38  & 28.88  & 32.27 & 30.01    & \textbf{32.99} & 31.70           \\
                        & 50\%                         & 25.27 & 21.82  & 25.44  & 27.99 & 28.57    & 30.23          & \textbf{30.47}  \\
                        & 70\%                         & 22.11 & 21.58  & 21.46  & 23.04 & 27.04    & 27.70          & \textbf{29.27}  \\ 
\midrule
\multirow{3}{*}{BSD300} & 30\%                         & 29.92 & 25.74  & 12.04  & 30.01 & 29.77    & 30.87          & \textbf{44.56}  \\
                        & 50\%                         & 26.32 & 24.50  & 6.01   & 26.25 & 26.19    & 27.84          & \textbf{38.03}  \\
                        & 70\%                         & 22.81 & 24.65  & 5.42   & 22.85 & 26.19    & 25.35          & \textbf{34.96}  \\
\bottomrule
\end{tabularx}
\end{table*}

In this section, the results of the training using the DIV2K Training Set dataset with noise levels of 30\%, 50\%, and 70\% will be presented along with an explanation. The test results using Lena, Bridge, Pepper, and BSD300 images with similar noise levels during training will be compared with the previous salt and pepper image denoising method.

\subsection{30\% Noise Level}

\begin{figure}[h]
    \centering
    \includegraphics[width=2.75in]{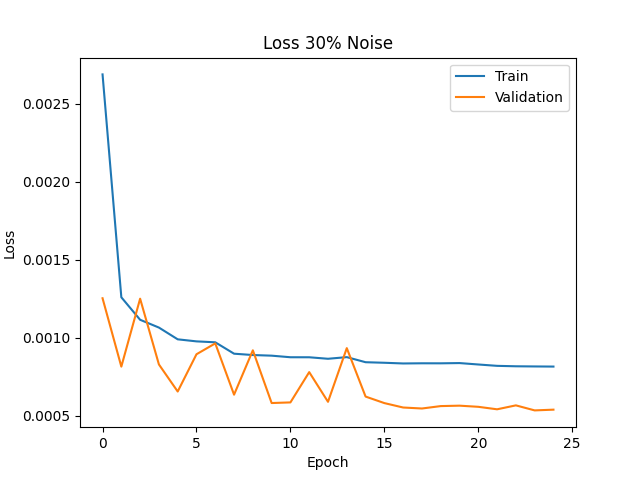}
    \caption{Training and Validation Loss Graph with 30\% Noise Level}
    \label{fig:loss30}
\end{figure}

Figure \ref{fig:loss30} depicts the training and validation loss graph from the training stage using a noise level of 30\%. The training was performed over the course of 40,000 iterations, separated into 25 epochs. The training loss value is 0.0026 in the first epoch and 0.0008 in the last epoch, whereas the validation loss value is 0.0012 in the first epoch and 0.0005 in the last epoch.

\begin{figure}[h]
    \centering
    \includegraphics[width=2.75in]{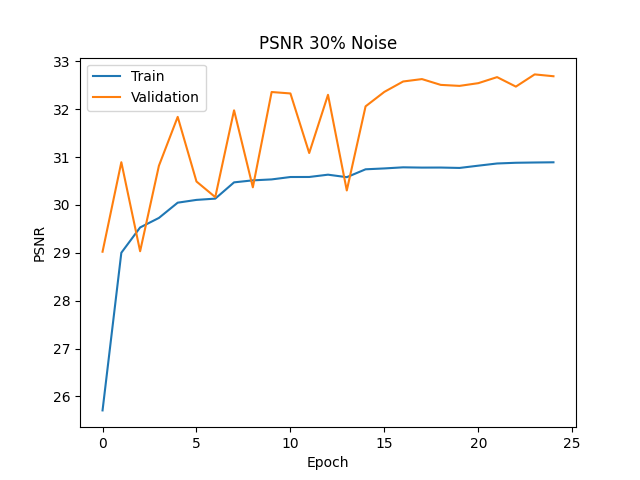}
    \caption{Training and Validation PSNR Graph with 30\% Noise Level}
    \label{fig:psnr30}
\end{figure}

Figure \ref{fig:psnr30} depicts the training and validation PSNR graph from the training stage using a noise level of 30\%. The training was performed over the course of 40,000 iterations, separated into 25 epochs. The training loss value is 25.707 in the first epoch and 30.891 in the last epoch, whereas the validation loss value is 29.022 in the first epoch and 32.689 in the last epoch.

\subsection{50\% Noise Level}

\begin{figure}[h]
    \centering
    \includegraphics[width=2.75in]{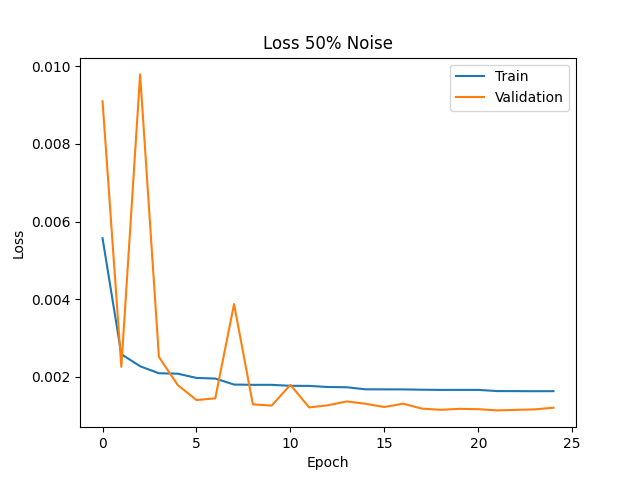}
    \caption{Training and Validation Loss Graph with 50\% Noise Level}
    \label{fig:loss50}
\end{figure}

Figure \ref{fig:loss50} depicts the training and validation loss graph from the training stage using a noise level of 50\%. The training was performed over the course of 40,000 iterations, separated into 25 epochs. The training loss value is 0.0055 in the first epoch and 0.0016 in the last epoch, whereas the validation loss value is 0.0090 in the first epoch and 0.0012 in the last epoch.

\begin{figure}[h]
    \centering
    \includegraphics[width=2.75in]{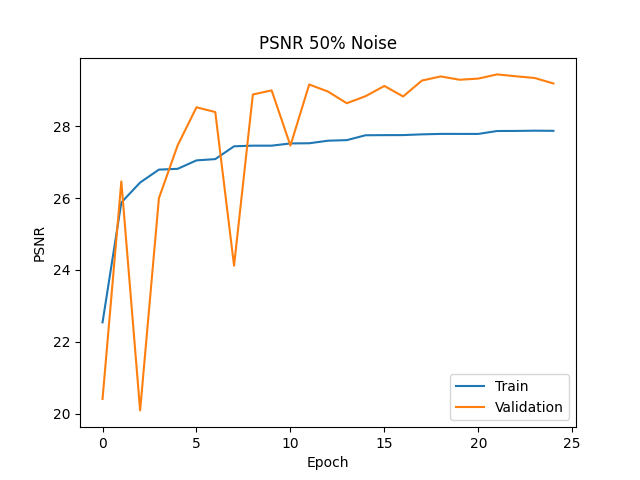}
    \caption{Training and Validation PSNR Graph with 50\% Noise Level}
    \label{fig:psnr50}
\end{figure}

Figure \ref{fig:psnr50} depicts the training and validation PSNR graph from the training stage using a noise level of 50\%. The training was performed over the course of 40,000 iterations, separated into 25 epochs. The training loss value is 22.541 in the first epoch and 27.868 in the last epoch, whereas the validation loss value is 20.410 in the first epoch and 29.185 in the last epoch.

\subsection{70\% Noise Level}

\begin{figure}[h]
    \centering
    \includegraphics[width=2.75in]{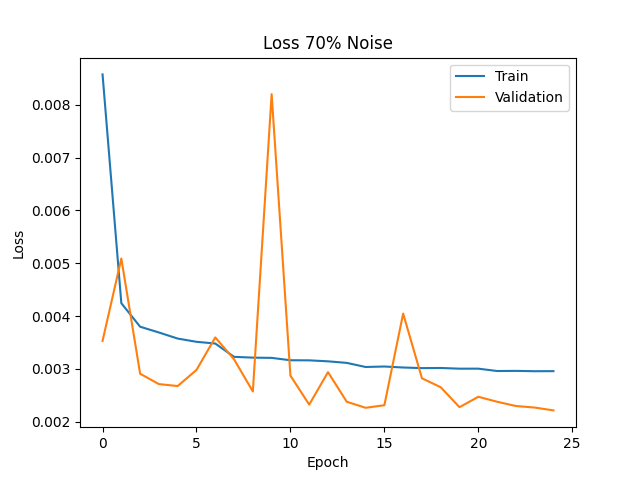}
    \caption{Training and Validation Loss Graph with 70\% Noise Level}
    \label{fig:loss70}
\end{figure}

Figure \ref{fig:loss70} depicts the training and validation loss graph from the training stage using a noise level of 70\%. The training was performed over the course of 40,000 iterations, separated into 25 epochs. The training loss value is 0.0085 in the first epoch and 0.0029 in the last epoch, whereas the validation loss value is 0.0035 in the first epoch and 0.0022 in the last epoch.

\begin{figure}[h]
    \centering
    \includegraphics[width=2.75in]{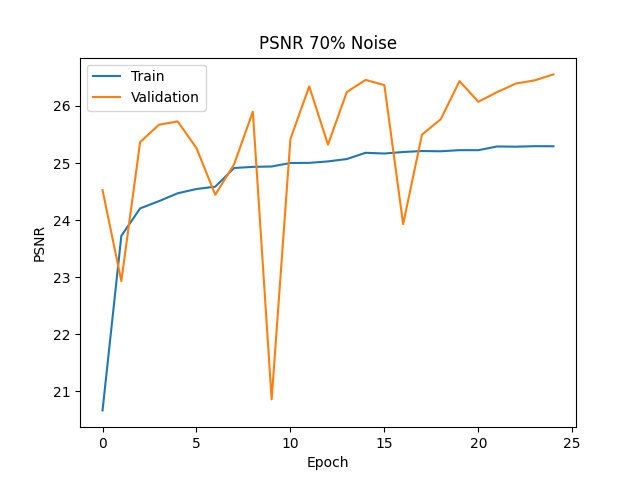}
    \caption{Training and Validation PSNR Graph with 70\% Noise Level}
    \label{fig:psnr70}
\end{figure}

Figure \ref{fig:psnr70} depicts the training and validation PSNR graph from the training stage using a noise level of 70\%. The training was performed over the course of 40,000 iterations, separated into 25 epochs. The training loss value is 20.665 in the first epoch and 25.294 in the last epoch, whereas the validation loss value is 24.524 in the first epoch and 26.553 in the last epoch.

\subsection{Testing}

\begin{figure*}[!t]
    \centering
    \subfloat[Ground Truth]{\label{fig:lena_a}\includegraphics[width=1.15in]{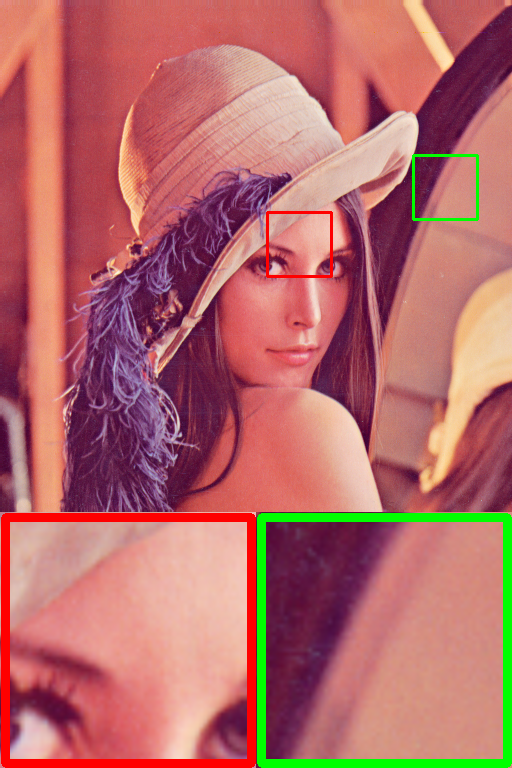}}
    \qquad
    \subfloat[30\% Noise]{\label{fig:lena_b30}\includegraphics[width=1.15in]{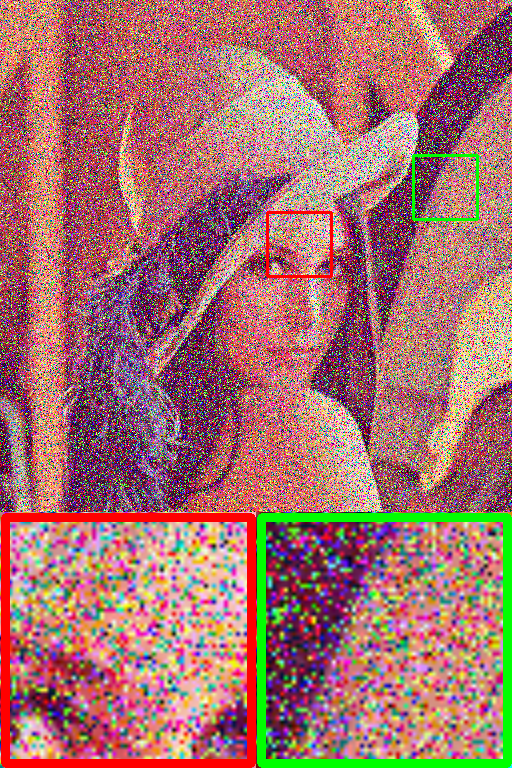}}
    \qquad
    \subfloat[30\% NSN]{\label{fig:lena_c30}\includegraphics[width=1.15in]{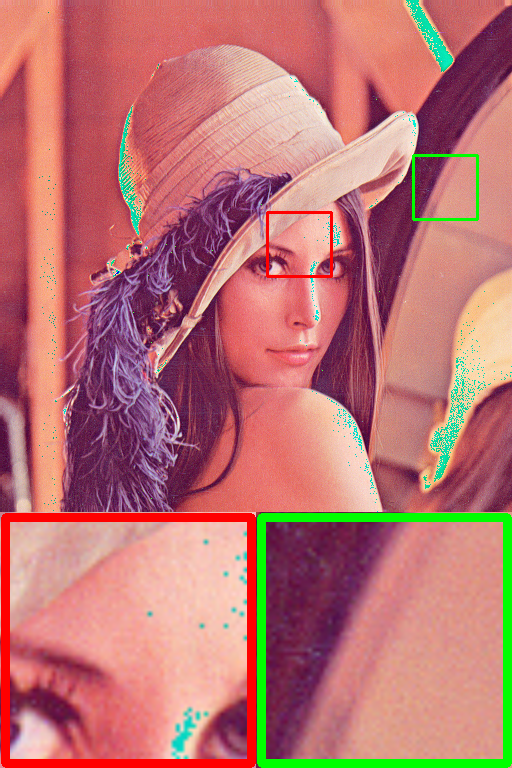}}
    \qquad
    \subfloat[50\% NSN]{\label{fig:lena_c50}\includegraphics[width=1.15in]{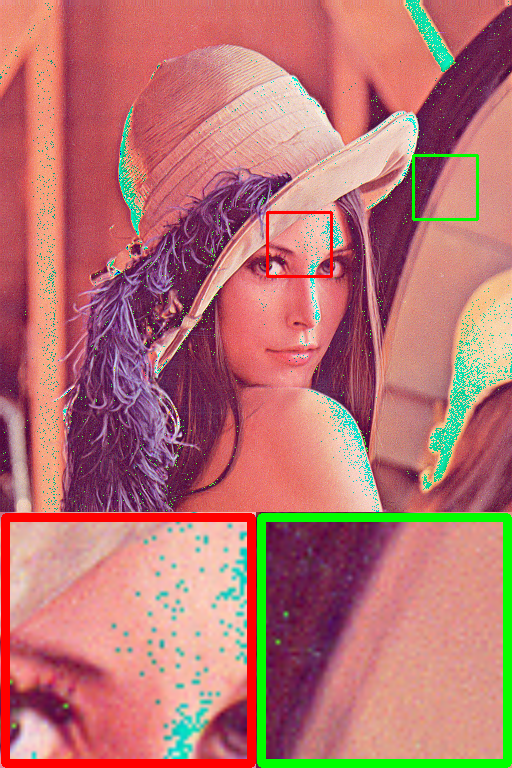}}
    \qquad
    \subfloat[70\% NSN]{\label{fig:lena_c70}\includegraphics[width=1.15in]{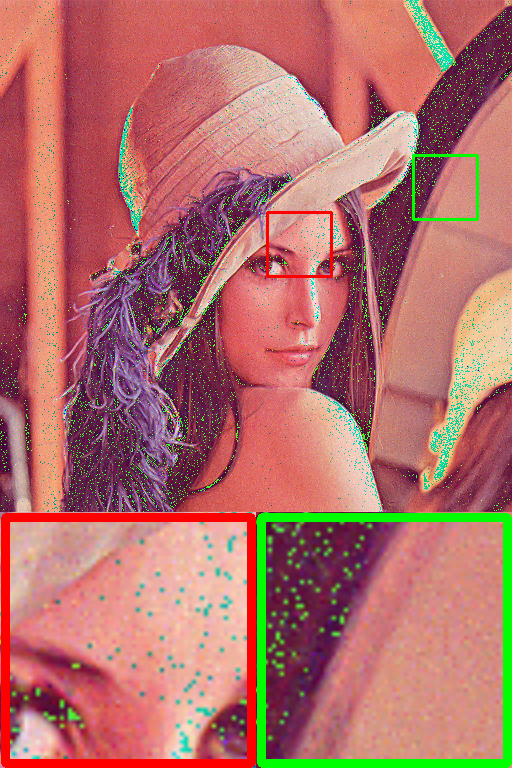}}
    \qquad
    \subfloat[50\% Noise]{\label{fig:lena_b50}\includegraphics[width=1.15in]{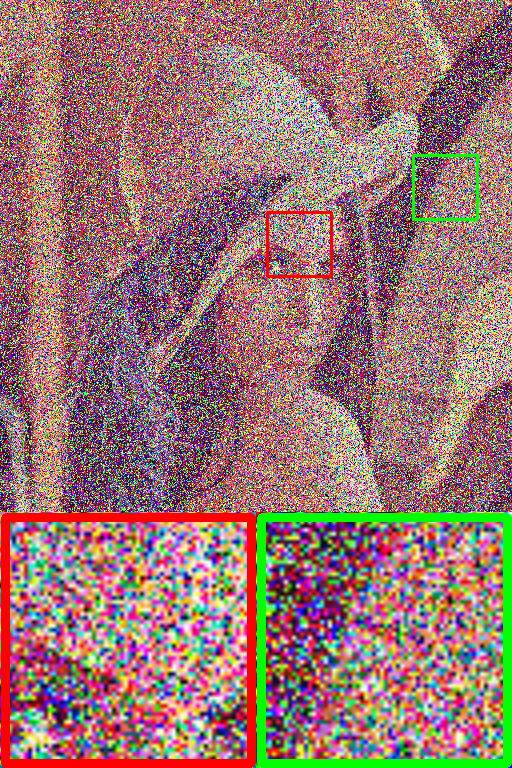}}
    \qquad
    \subfloat[70\% Noise]{\label{fig:lena_b70}\includegraphics[width=1.15in]{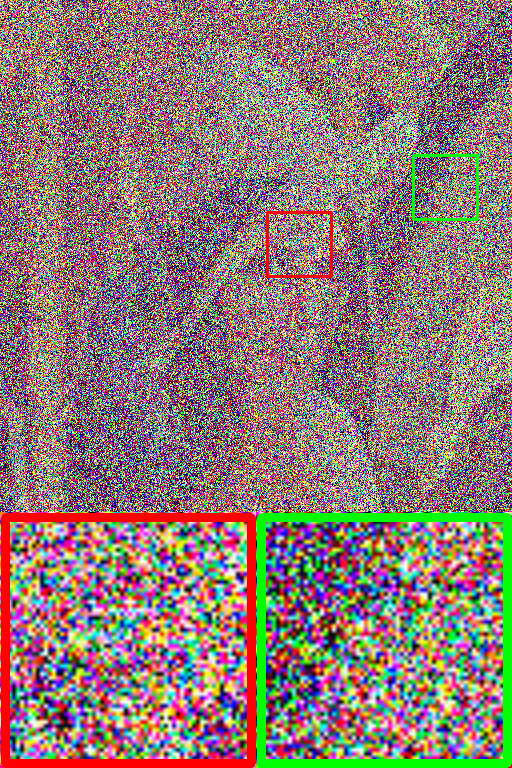}}
    \qquad
    \subfloat[30\% SEN]{\label{fig:lena_d30}\includegraphics[width=1.15in]{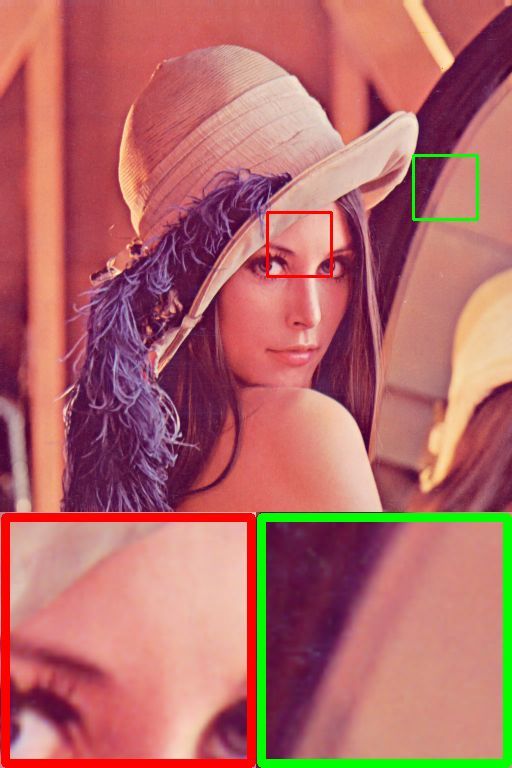}}
    \qquad
    \subfloat[50\% SEN]{\label{fig:lena_d50}\includegraphics[width=1.15in]{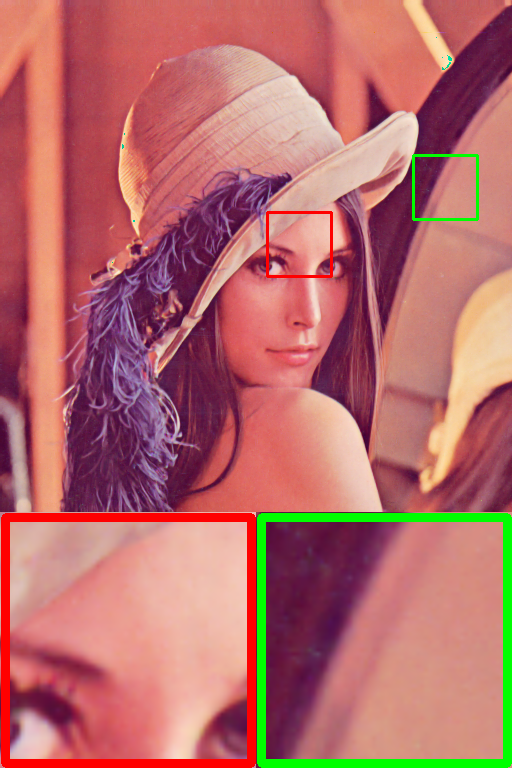}}
    \qquad
    \subfloat[70\% SEN]{\label{fig:lena_d70}\includegraphics[width=1.15in]{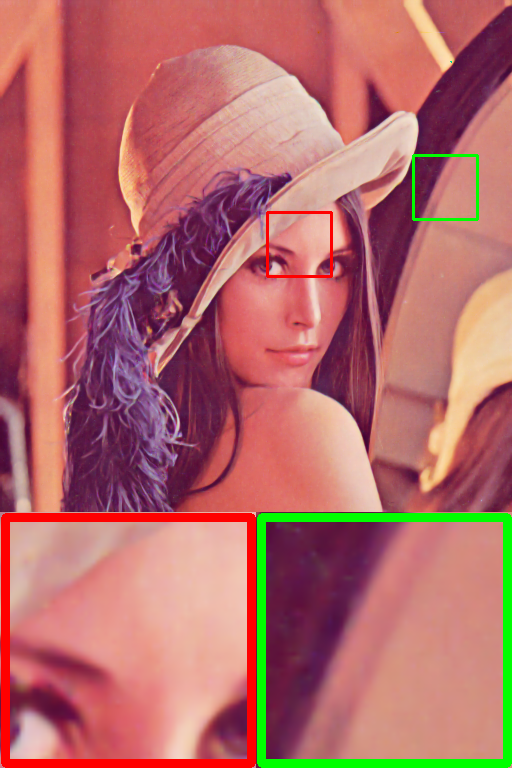}}
    \caption{Visual investigation of denoised "Lena" image from the proposed method over various noise levels $p=\{0.3,0.5,0.7\}$}
    \label{fig:lena}
\end{figure*}

\begin{figure*}[!t]
    \centering
    \subfloat[Ground Truth]{\label{fig:bridge_a}\includegraphics[width=1.15in]{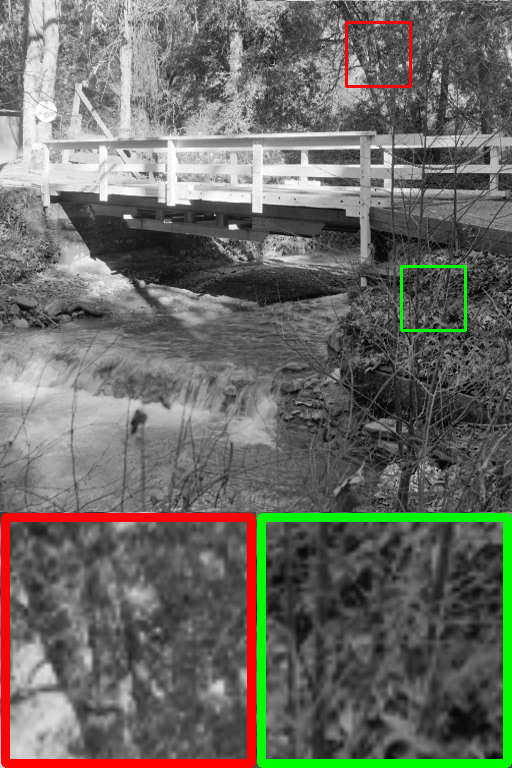}}
    \qquad
    \subfloat[30\% Noise]{\label{fig:bridge_b30}\includegraphics[width=1.15in]{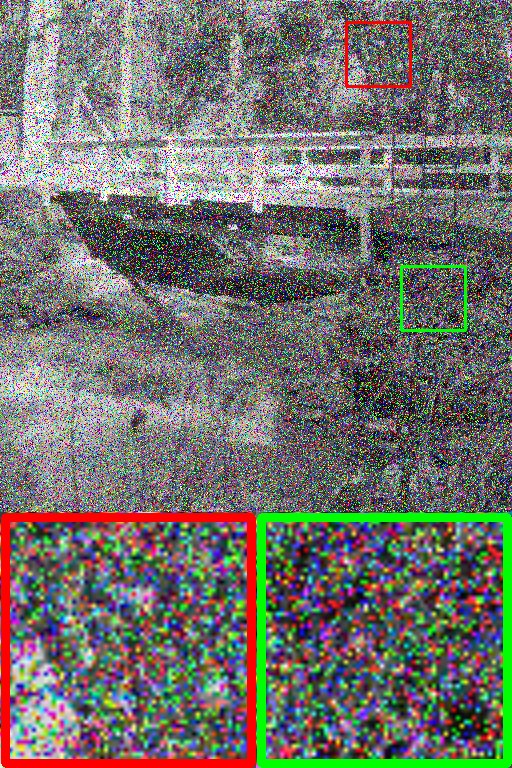}}
    \qquad
    \subfloat[30\% NSN]{\label{fig:bridge_c30}\includegraphics[width=1.15in]{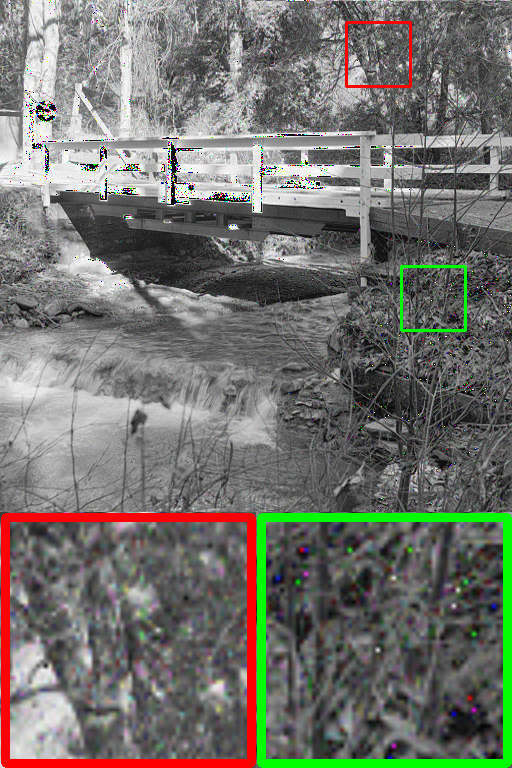}}
    \qquad
    \subfloat[50\% NSN]{\label{fig:bridge_c50}\includegraphics[width=1.15in]{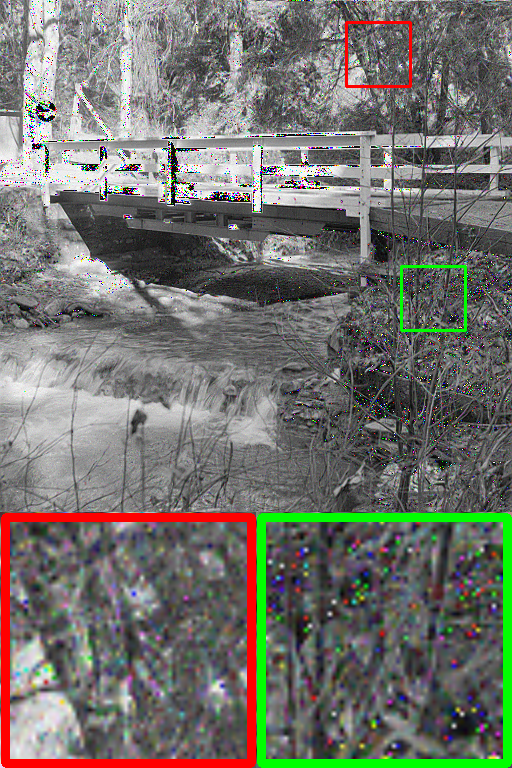}}
    \qquad
    \subfloat[70\% NSN]{\label{fig:bridge_c70}\includegraphics[width=1.15in]{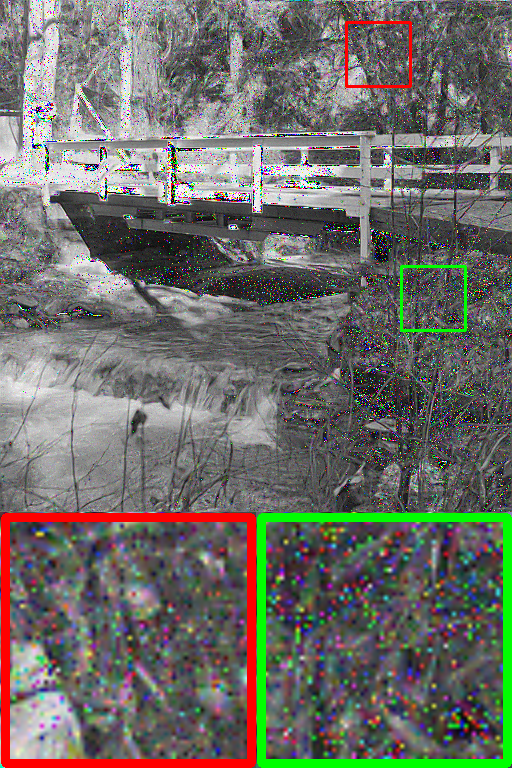}}
    \qquad
    \subfloat[50\% Noise]{\label{fig:bridge_b50}\includegraphics[width=1.15in]{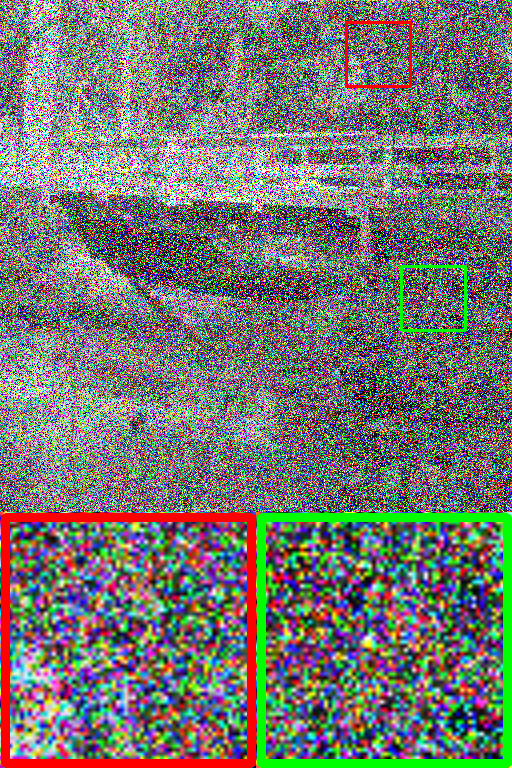}}
    \qquad
    \subfloat[70\% Noise]{\label{fig:bridge_b70}\includegraphics[width=1.15in]{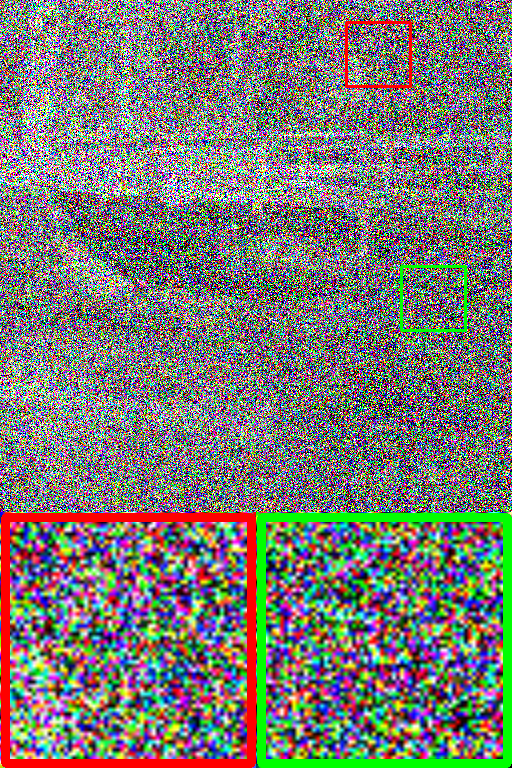}}
    \qquad
    \subfloat[30\% SEN]{\label{fig:bridge_d30}\includegraphics[width=1.15in]{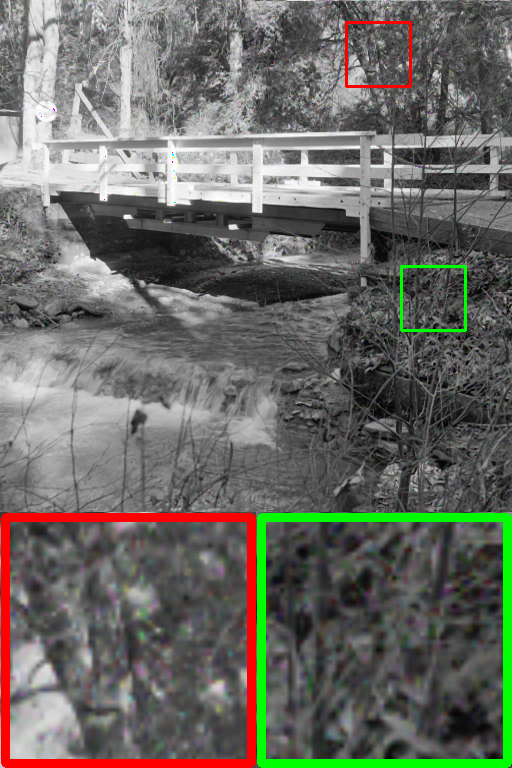}}
    \qquad
    \subfloat[50\% SEN]{\label{fig:bridge_d50}\includegraphics[width=1.15in]{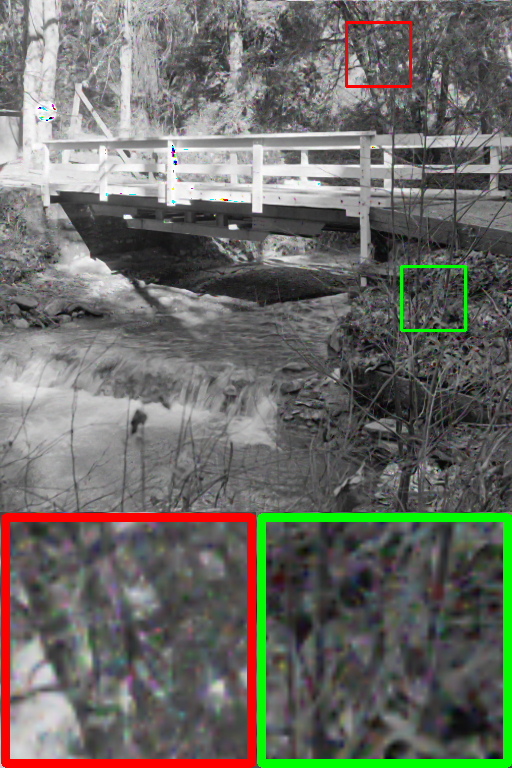}}
    \qquad
    \subfloat[70\% SEN]{\label{fig:bridge_d70}\includegraphics[width=1.15in]{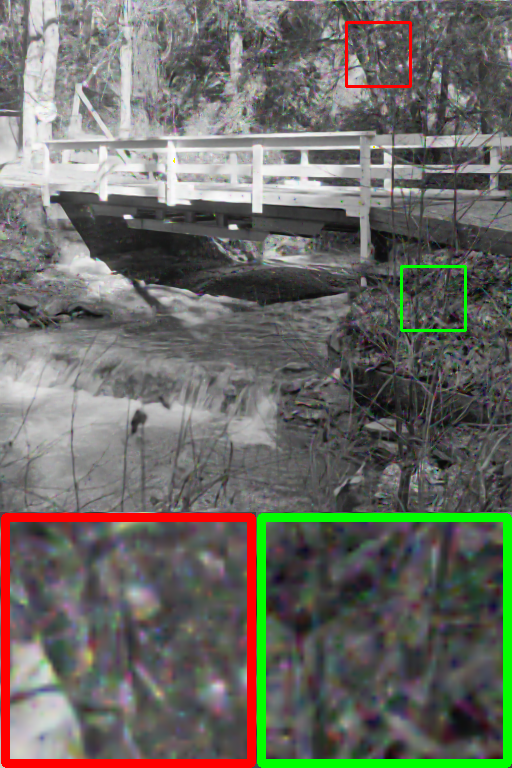}}
    \caption{Visual investigation of denoised "Bridge" image from the proposed method over various noise levels $p=\{0.3,0.5,0.7\}$}
    \label{fig:bridge}
\end{figure*}

\begin{figure*}[!t]
    \centering
    \subfloat[Ground Truth]{\label{fig:pepper_a}\includegraphics[width=1.15in]{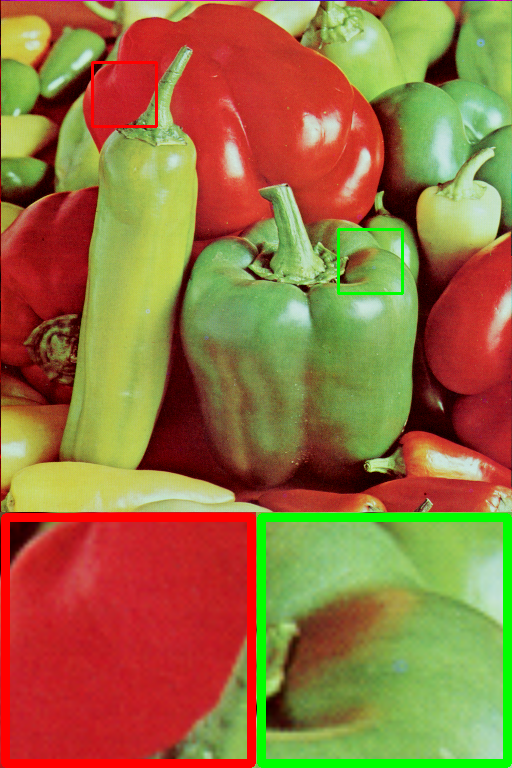}}
    \qquad
    \subfloat[30\% Noise]{\label{fig:pepper_b30}\includegraphics[width=1.15in]{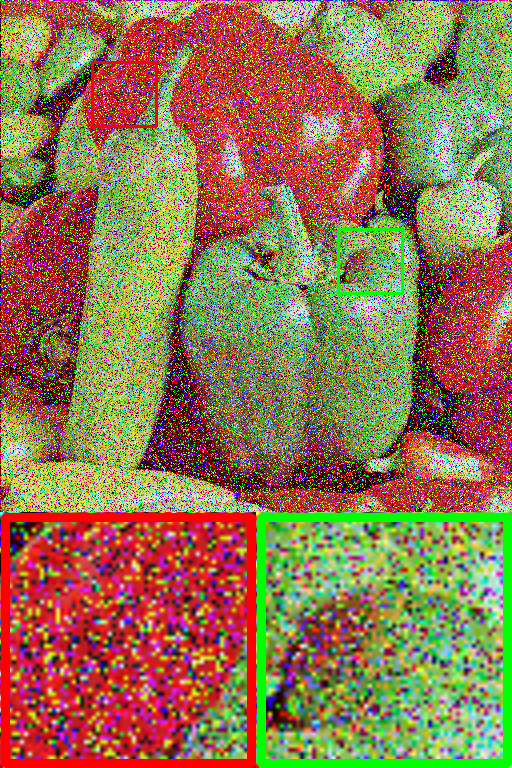}}
    \qquad
    \subfloat[30\% NSN]{\label{fig:pepper_c30}\includegraphics[width=1.15in]{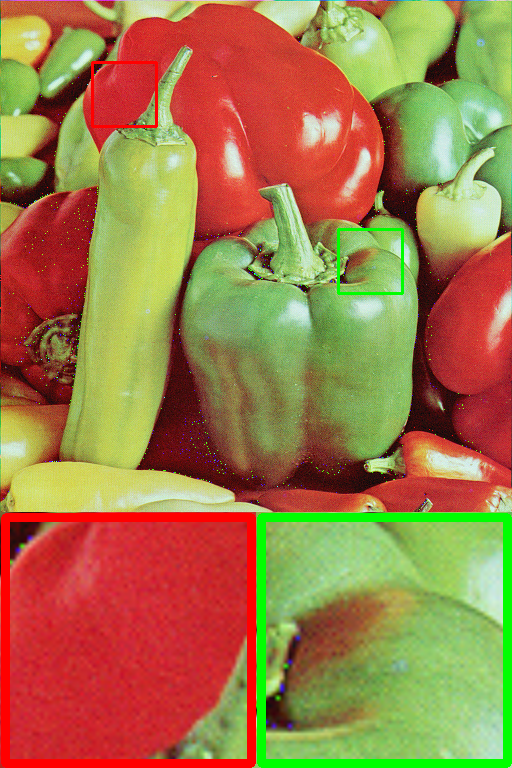}}
    \qquad
    \subfloat[50\% NSN]{\label{fig:pepper_c50}\includegraphics[width=1.15in]{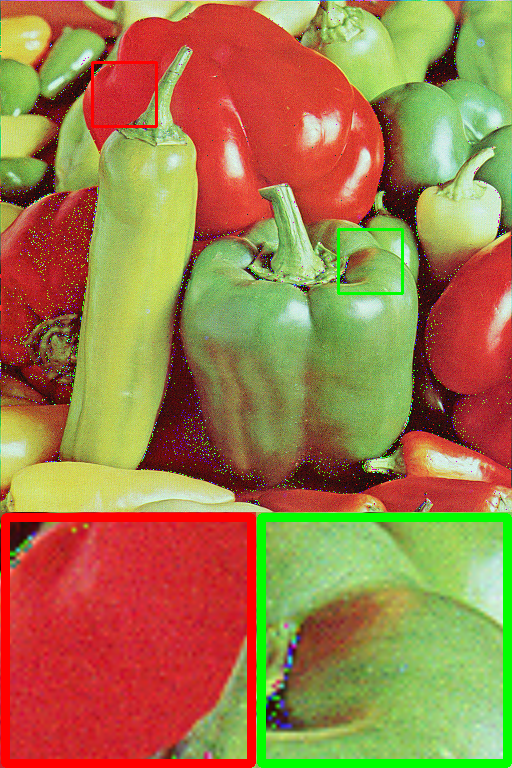}}
    \qquad
    \subfloat[70\% NSN]{\label{fig:pepper_c70}\includegraphics[width=1.15in]{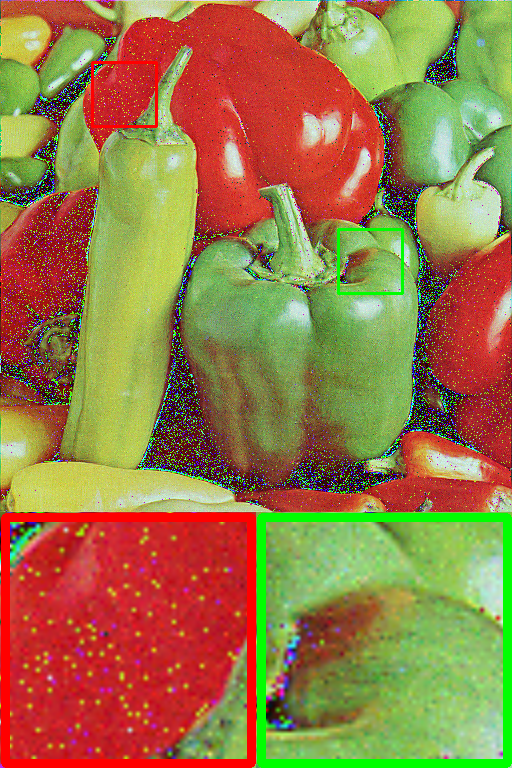}}
    \qquad
    \subfloat[50\% Noise]{\label{fig:pepper_b50}\includegraphics[width=1.15in]{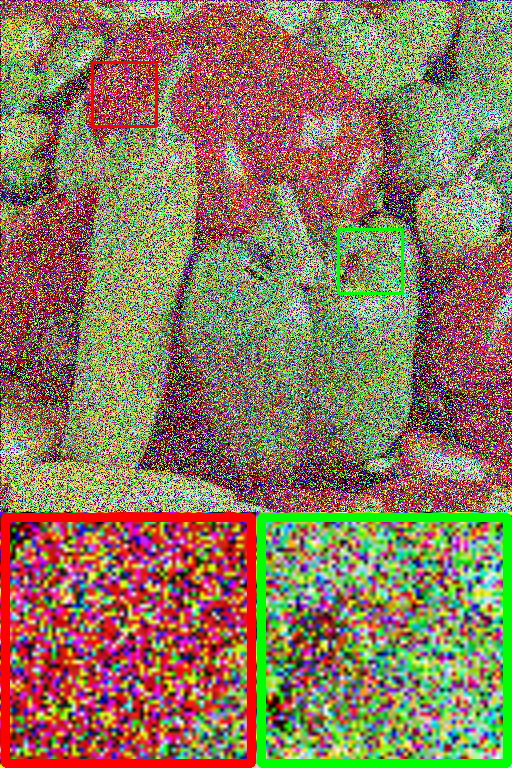}}
    \qquad
    \subfloat[70\% Noise]{\label{fig:pepper_b70}\includegraphics[width=1.15in]{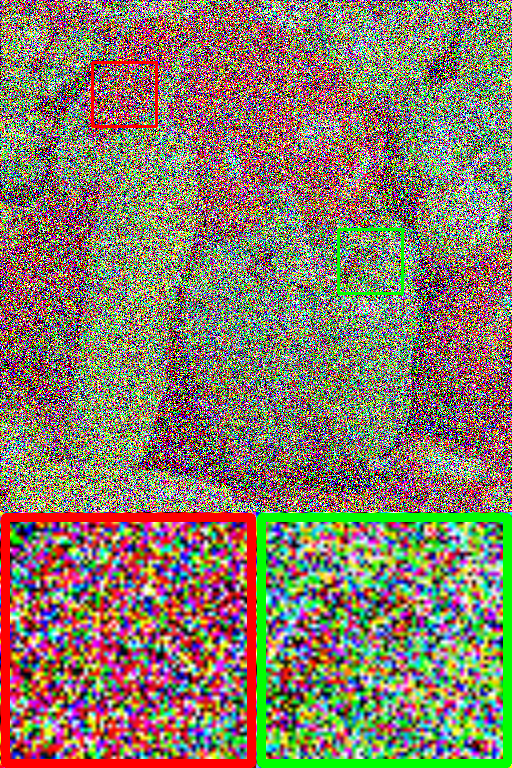}}
    \qquad
    \subfloat[30\% SEN]{\label{fig:pepper_d30}\includegraphics[width=1.15in]{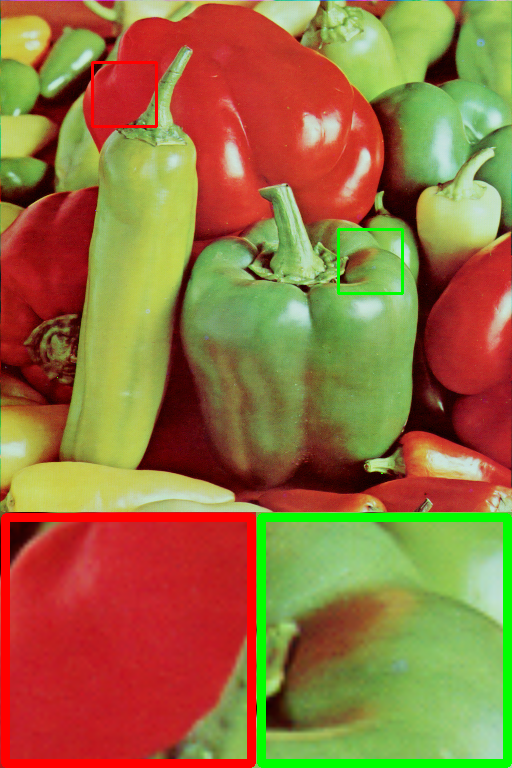}}
    \qquad
    \subfloat[50\% SEN]{\label{fig:pepper_d50}\includegraphics[width=1.15in]{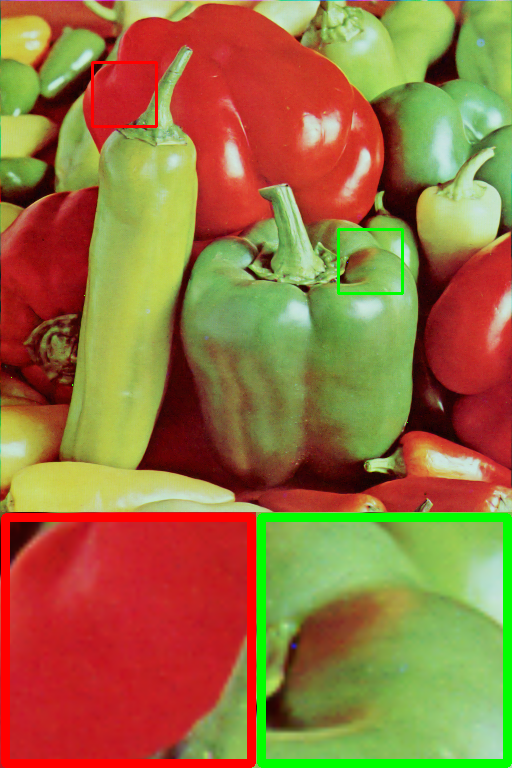}}
    \qquad
    \subfloat[70\% SEN]{\label{fig:pepper_d70}\includegraphics[width=1.15in]{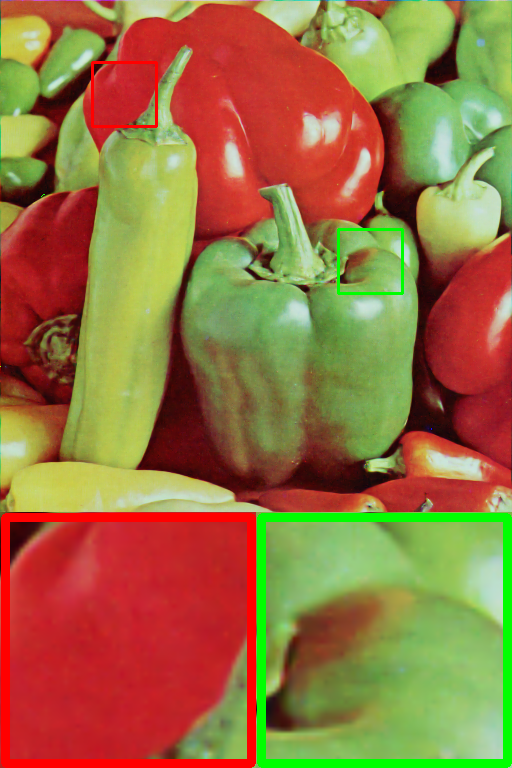}}
    \caption{Visual investigation of denoised "Pepper" image from the proposed method over various noise levels $p=\{0.3,0.5,0.7\}$}
    \label{fig:pepper}
\end{figure*}

After the best hyperparameters were found, we tested the model using  Lena, Bridge, Pepper, and BSD300 images with the noise levels for each image of 30\%, 50\%, and 70\%, respectively. After testing, the PSNR values were compared with the previous method, namely DBA \cite{srinivasan2007dba}, NASNLM \cite{varghese2015nasnlm}, PARIGI \cite{delon2016parigi}, NLSF \cite{fu2018nlsf}, NLSF-MLP \cite{burger2012nlsf}, and NLSF-CNN \cite{fu2018nlsf}. Table \ref{tab:result} shows that the proposed method excels in all cases except for Pepper's image with a noise level of 30\%, where NLSF-CNN excels with a PSNR value of 32.99 while the proposed method gets a PSNR value of 31.70. A visual comparison of the test results is presented in Figure \ref{fig:lena} for the Lena image, Figure \ref{fig:bridge} for the Bridge image, and Figure \ref{fig:pepper} for the Pepper image.

\subsection{Discussion}

\begin{figure}[!t]
    \centering
    \subfloat[Ground Truth]{\label{fig:blur_a}\includegraphics[width=1in]{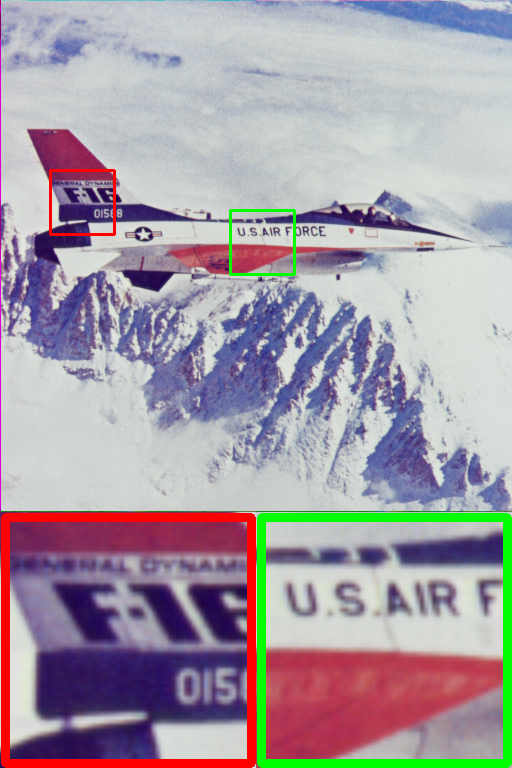}}
    \quad
    \subfloat[Blurry]{\label{fig:blur_b}\includegraphics[width=1in]{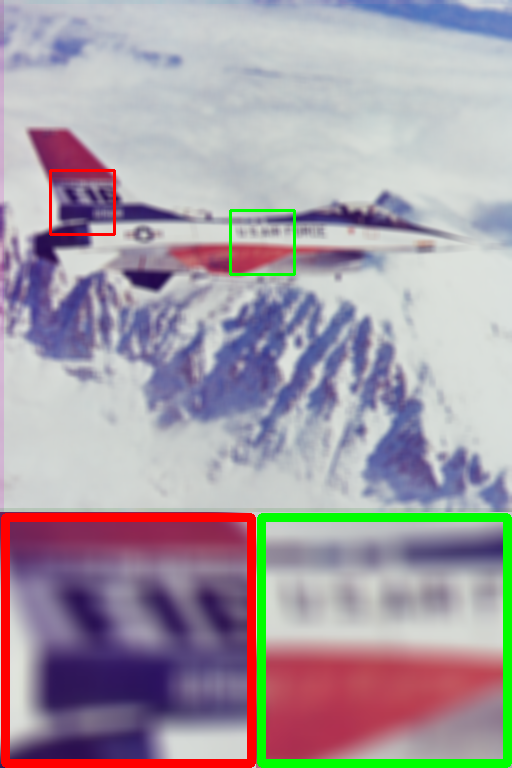}}
    \quad
    \subfloat[Deblurred]{\label{fig:blur_c}\includegraphics[width=1in]{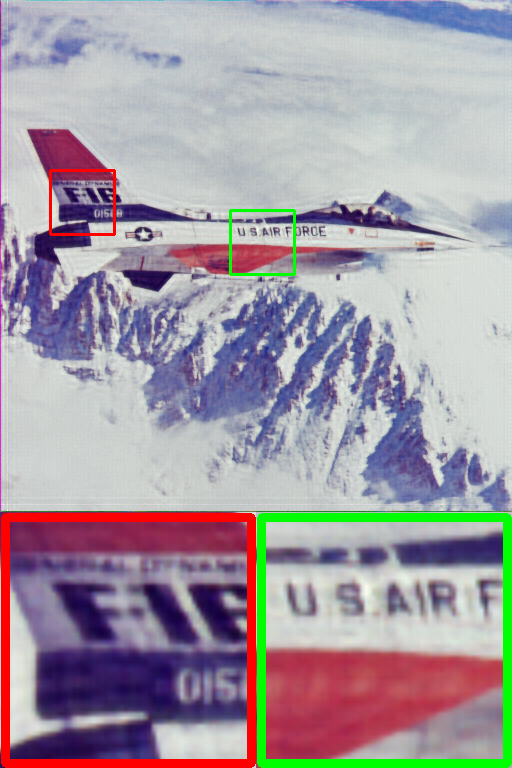}}
    \caption{Visual investigation of deblurred "Airplane" image using the proposed method over $7\times7$ blur kernel}
    \label{fig:airplane}
\end{figure}

\begin{figure}[!t]
    \centering
    \subfloat[Ground Truth]{\label{fig:jpeg_a}\includegraphics[width=1in]{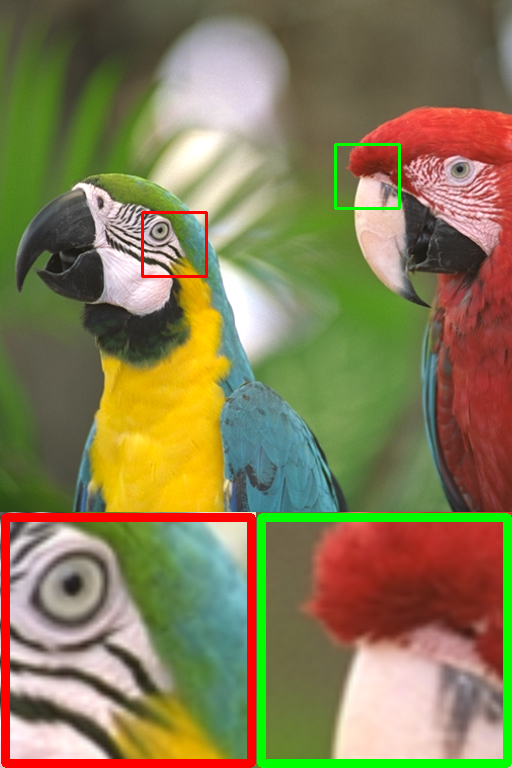}}
    \quad
    \subfloat[Artifact]{\label{fig:jpeg_b}\includegraphics[width=1in]{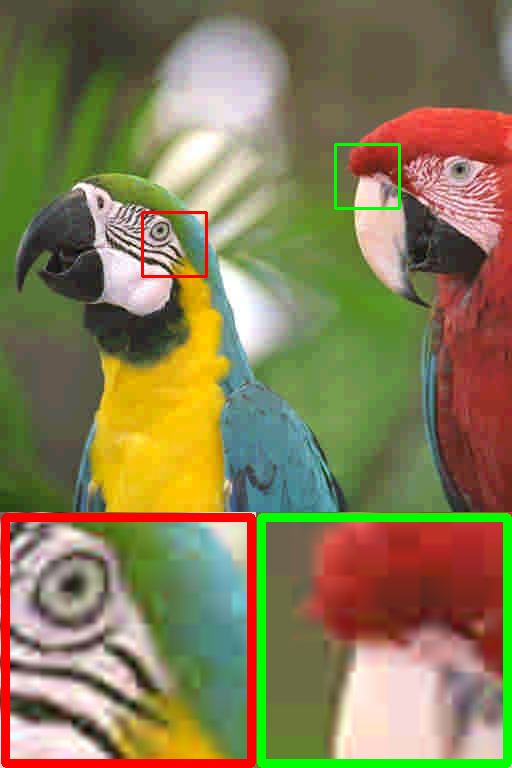}}
    \quad
    \subfloat[Cleaned]{\label{fig:jpeg_c}\includegraphics[width=1in]{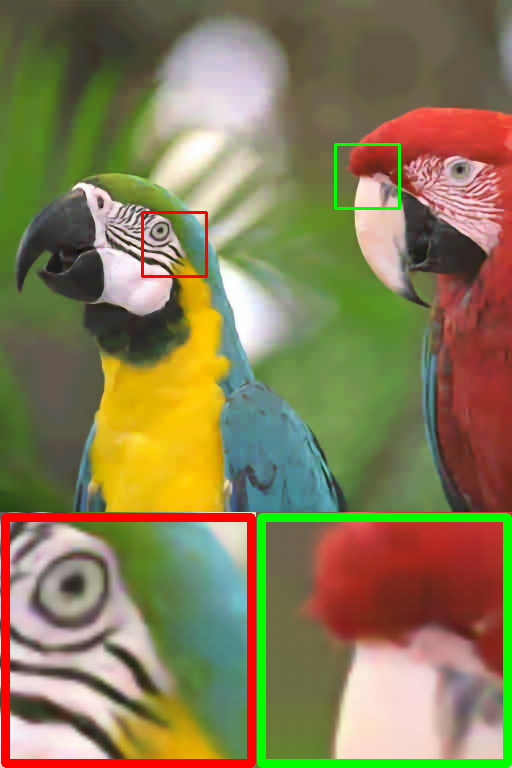}}
    \caption{Visual investigation of JPEG artifact removed "Parrot" image using the proposed method over 10\% JPEG compression}
    \label{fig:parrot}
\end{figure}

\begin{figure}[!t]
    \centering
    \subfloat[Ground Truth]{\label{fig:sisr_a}\includegraphics[width=1in]{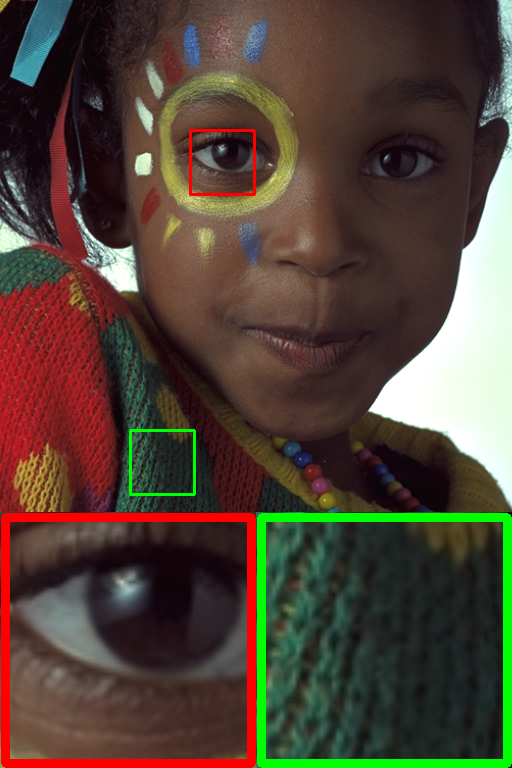}}
    \quad
    \subfloat[Low Res]{\label{fig:sisr_b}\includegraphics[width=1in]{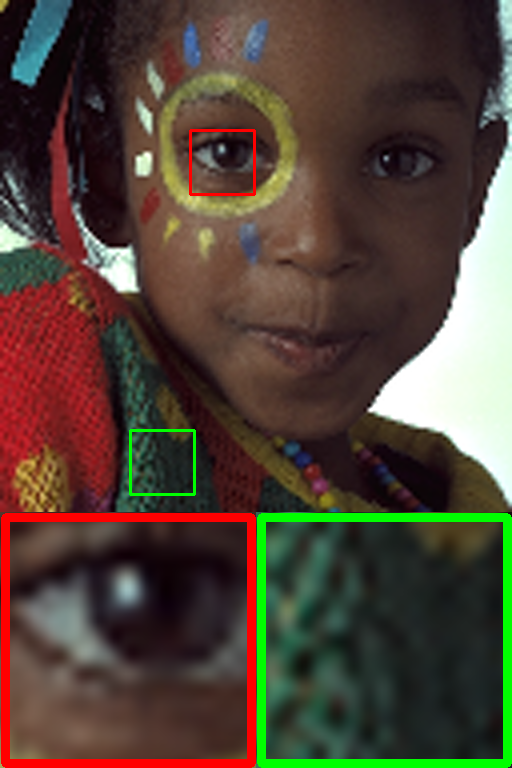}}
    \quad
    \subfloat[Upscaled]{\label{fig:sisr_c}\includegraphics[width=1in]{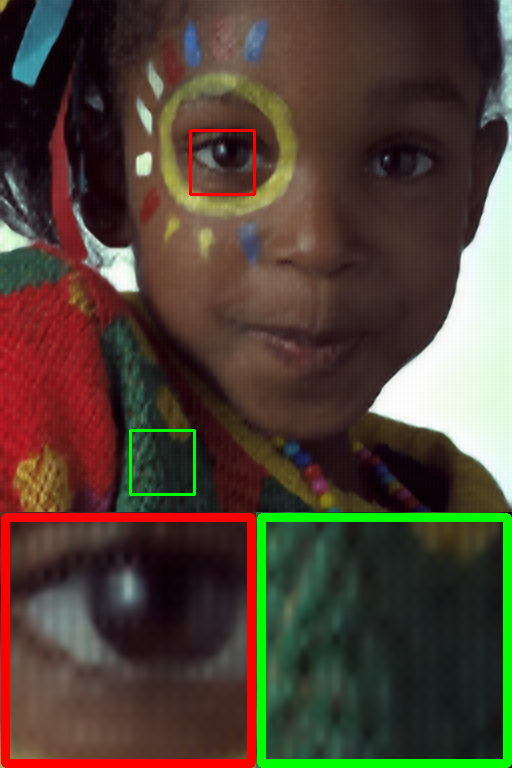}}
    \caption{Visual investigation of Single Image Super Resolution "Girl" image using the proposed method over $4\times$ bicubic downsampled image}
    \label{fig:girl}
\end{figure}

In this section, we will explore the potential of the Residual Transformer Fusion Network architecture in solving other cases such as image deblurring, JPEG artifact removal, and Single Image Super-Resolution (SISR). The model was trained using the same parameters except for epochs, which in this experiment, we only used 10 epochs.

In Figure \ref{fig:airplane}, the deblurred image shows sharp results with great details, although it still leaves a shadow image in the perimeter of contrasting region. In Figure \ref{fig:parrot}, the JPEG artifact removal image shows a smoother result when the color transition occurs, but the fine details cannot be reconstructed. Figure \ref{fig:girl} shows the results of the Single Image Super-Resolution which produces a sharp and contrasting image but still leaves an artifact in the form of a grid pattern.

\section{Conclusions}
\label{sec:conc}

This study aims to obtain a neural network architecture to reconstruct noisy images with salt and pepper noise types effectively. In this study, a neural network architecture called Residual Transformer Fusion Network is proposed, which is a combination of Residual Network (ResNet) which is used as the Noise Suppression Network (NSN), and Convolutional Vision Transformer (CvT) which is used in the Structure Enhancement Network (SEN). The proposed architecture is trained with 40,000 iterations which are divided into 25 epochs using the DIV2K Training Set dataset which has been applied to the patching process with a patch size of 64×64 and validation using the DIV2K Validation Set dataset which is not applied by the patching process. From the test results using Lena, Bridge, Pepper, and BSD300 images with noise levels of 30\%, 50\%, and 70\%, the proposed method provides a better PSNR than the previous methods, namely DBA \cite{srinivasan2007dba}, NASNLM \cite{varghese2015nasnlm}, PARIGI \cite{delon2016parigi}, NLSF \cite{fu2018nlsf}, NLSF-MLP \cite{burger2012nlsf}, dan NLSF-CNN \cite{fu2018nlsf}. The visual comparison shows that the proposed method can produce clean images with good detail.

\subsection*{Acknowledgements}

This research was carried out with the support of the computational facilities of the Intelligent Systems and Humanized Computing Laboratory of Sebelas Maret University Informatics.

\nocite{}
\bibliography{references}
\bibliographystyle{IEEEtran}

\end{document}